\documentclass[conference]{IEEEtran}
\IEEEoverridecommandlockouts

\usepackage{cite}
\usepackage{amsmath,amssymb,amsfonts}
\usepackage{array}
\usepackage{booktabs}
\usepackage{graphicx}
\usepackage{makecell}
\usepackage{multirow}
\usepackage{pifont}
\usepackage{subfigure}
\usepackage{textcomp}
\usepackage{xcolor}
\usepackage{url}
\usepackage[hidelinks]{hyperref}

\begin{document}

\title{Toward General Digraph Contrastive Learning: \\A Dual Spatial Perspective}

\author{Zhengyu Wu, Daohan Su,Yang Zhang, Xunkai Li, Rong-Hua Li, Guoren Wang}

\maketitle

\begin{abstract}
Graph Contrastive Learning (GCL) has emerged as a powerful tool for extracting consistent representations from graphs, independent of labeled information.
However, existing methods predominantly focus on undirected graphs, disregarding the pivotal directional information that is fundamental and indispensable in real-world networks (e.g., social networks and recommendations).
In this paper, we introduce S2-DiGCL, a directed graph (digraph) contrastive learning framework that jointly models directional information from the complex and real domains.
From the complex-domain perspective, S2-DiGCL introduces personalized perturbations into the magnetic Laplacian to adaptively modulate edge phases and directional semantics.
From the real-domain perspective, it employs a path-based subgraph augmentation strategy to capture fine-grained local asymmetries and topological dependencies.
By jointly leveraging these two complementary spatial views, S2-DiGCL learns direction-aware representations for self-supervised digraph learning.
Experiments on seven real-world digraph datasets show that S2-DiGCL consistently improves node classification performance over strong unsupervised baselines, with an average gain of 4.41\%, and also achieves strong link prediction results on four benchmark datasets.
\end{abstract}

\begin{IEEEkeywords}
Graphs and networks, Directed graph, Machine learning, Contrastive learning.
\end{IEEEkeywords}

\section{Introduction}

Graph has become a fundamental data structure for modeling pairwise relationships across diverse domains, such as social interaction~\cite{mislove2007measurement}, behavior analysis~\cite{Human_Action}, networks~\cite{lin2013complex, haznagy2015complex}, and recommendation systems~\cite{su2024dcl}.
This widespread use has spurred the rapid development of GNNs~\cite{reiser2022graph, wu2020gnn_survey1}, which effectively capture topological dependencies and node interactions.
Despite advancements, conventional supervised GNNs face inherent limitations due to their reliance on extensive labeled data, posing a critical bottleneck as the volume of real-world graphs continues to grow while annotated data remain scarce and expensive to obtain.
To mitigate this limitation, GCL~\cite{velickovic2018dgi, zhu2020deep} has emerged as a promising self-supervised paradigm that learns robust and transferable node representations by enforcing consistency across multiple augmented graph views.

\textbf{Observation.}
While current GCL methods have demonstrated strong performance on undirected graphs, their applicability to digraphs remains largely unexplored.
The persistent neglect of directional semantics remains a major challenge for extending GCL to digraph settings.
Owing to the unique advantages and fundamental role of edge directionality in guiding information propagation, digraphs motivate us to engage in an in-depth exploration in contrastive learning.

\begin{figure}[t]
\centering
\includegraphics[width=\linewidth]{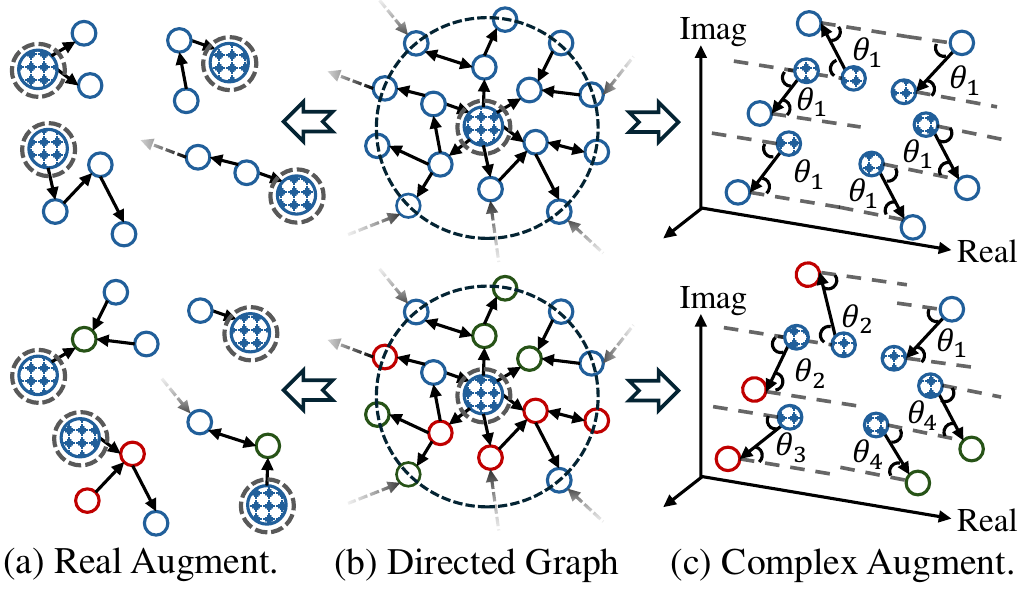}
\vspace{-0.5cm}
\caption{Illustration of naive (Upper) vs. personalized (Lower) augmentations from real and complex spatial perspectives.
Uniform perturbations fail to distinguish heterophilous nodes or asymmetric directional flows, while personalized perturbations adapt augmentation strength and phase shifts to each node’s topology, enabling direction-aware contrastive learning in S2-DiGCL.}
\vspace{-0.5cm}
\label{fig.motivation}
\end{figure}

\textbf{Motivation.}
\textit{Digraphs inherently encode directional information flows between entities, providing a more accurate representation of real-world scenarios.}
For example, directed edges capture asymmetric influence in social networks\cite{gupte2011finding}, one-way constraints and congestion in transportation systems\cite{marshall2018street}, and unidirectional relationships in citation networks~\cite{hummon1989connectivity}.
Beyond these intuitive examples, recent studies~\cite{maekawa2023a2dug,dirgnn_rossi_2023,sun2023adpa } have revealed a crucial insight:
\textit{The direction of edge offers a novel perspective for addressing the persistent challenges of topological heterophily}, which refers to connected nodes exhibiting dissimilar features.
Collectively, these findings highlight the untapped potential of directional information and motivate the development of a dedicated contrastive learning framework for digraphs.

To achieve effective digraph contrastive learning, a straightforward approach is to directly apply augmentation strategies designed for undirected graphs (e.g., random edge dropping or feature masking).
However, due to the intrinsic structural asymmetry of directed graphs, such approaches often disrupt directional information flows patterns and consequently yield suboptimal performance.
Although several recent investigations into digraph contrastive learning~\cite{tong2021directed, ko2023universal} have shown promise, they remain hindered by two notable \textbf{Limitations.}\\
\ding{192} \textit{Singular Perspective.}
Current methods restrict their augmentation strategies to either real-valued or complex-valued domain.
Perturbations in the real domain explicitly capture local directional dependencies, whereas those in the complex domain implicitly encode global rotational semantics via the imaginary components of magnetic Laplacians~\cite{zhang2021magnet}.
However, relying solely on either perspective is insufficient to comprehensively capture the multifaceted characteristics of digraphs.\\
\ding{193} \textit{Lack of Personalization.}
Current perturbation strategies employ uniform augmentation parameters for all nodes, overlooking the inherent diversity in directed topologies and complex semantic contexts unique to each node.
For instance, in Fig.\ref{fig.motivation}, hub nodes in citation networks may require different augmentation intensities compared to periphery nodes, owing to their distinct in-degree and out-degree distributions.
Such one-size-fits-all strategies hinder the generation of context-aware embeddings by ignoring node-specific characteristics.
These limitations motivate us to advance digraph contrastive learning by designing frameworks that jointly exploit both local and global directional information while incorporating personalized augmentation to adapt to diverse node contexts.

\textbf{Solution.}
To address these challenges, we propose \textbf{S2-DiGCL} (Dual \textbf{S}patial \textbf{Di}rected \textbf{G}raph \textbf{C}ontrastive \textbf{L}earning), where `S2' signifies the dual spatial perspectives that jointly model directional information in real and complex domains.
S2-DiGCL improves self-supervised representation learning on digraphs through two complementary design choices:
\ding{182} \textbf{Comprehensive Perspectives.}
We develop a path-based augmentation strategy in the real domain to preserve local directional neighborhoods through biased random walks.
It is then complemented by a complex-domain magnetic augmentation mechanism that captures global directional dependencies through the imaginary-space modulation of the magnetic Laplacian.
Together, these dual perturbations help construct highly complementary views that encode both local directional interactions and global rotational semantics, acquiring a more complete representation of digraph structures.
\ding{183} \textbf{Personalized Augmentation.}
We further introduce a personalized magnetic Laplacian augmentation technique, which perturbs both the Laplacian matrix and the digraph structure.
This personalization accommodates a diverse range of global-structured patterns and ensures that augmentations align more closely with each node’s unique context.
We also adapt the traditional random walk approach to the unique topology of digraphs, ensuring that the walking process generates personalized perturbations based on the digraph’s specific structural properties.
By integrating these dual spatial insights, S2-DiGCL produces complementary contrastive views for digraph learning.

\textbf{Our contributions.}
Building upon these core ideas, our work makes the following three key advancements:
(1) \underline{\textit{General Framework.}}
We propose S2-DiGCL, a dual-spatial contrastive learning framework for digraphs that jointly leverages perturbations in both the real and complex domains to maximize the modeling of directional information.
(2) \underline{\textit{Innovative Perturbation Techniques.}}
We devise a path-based real-domain augmentation to generate local context-level views, complemented by a personalized complex-domain magnetic perturbation that captures global directional dependencies, together enabling a more comprehensive representation of digraph structures.
(3) \underline{\textit{Empirical Effectiveness.}}
Experiments on seven real-world digraph datasets show that S2-DiGCL consistently improves node classification performance over strong unsupervised baselines, and also achieves competitive link prediction results on four benchmark datasets.

\section{Preliminaries}
\subsection{Notation}
Let $\mathcal{G} = (\mathcal{V}, \mathcal{E})$ represent a directed graph, where $\mathcal{V}$ and $\mathcal{E}$ represent the sets of nodes and edges, respectively, with $\mathcal{|V|} = n$ and $\mathcal{|E|} = m$.
Each node $v\in \mathcal{V}$ is associated with a $d$-dimensional feature vector, and all node features are stacked into a feature matrix $\mathbf{X}\in\mathbb{R}^{n\times d}$.
The topology of directed graph $\mathcal{G}$ is represented by an asymmetrical adjacency matrix $\mathbf{A}=\{0,1\}^{n\times n}$ that encodes the direction of edges between nodes.
Specifically, for any pair of nodes $u,v\in\mathcal{V}$, $\mathbf{A}(u,v)=1$ if $(u,v)\in\mathcal{E}$, and $\mathbf{A}(u,v)=0$ vice versa.

\subsection{Graph Contrastive Learning}
Graph Contrastive Learning (GCL) has emerged as an effective paradigm for self-supervised graph representation learning, alleviating the dependence on labeled data.
Recent surveys provide a comprehensive overview of its objectives, augmentation strategies, and open challenges~\cite{ju2024gclsurvey}.
GCL leverages the intrinsic topological structure of graphs to enforce representation consistency across multiple augmented views without reliance on labels, thereby pulling positive samples closer and pushing negative samples apart in the embedding space.
Depending on the granularity of contrastive objectives, existing GCL methods can be categorized into intra-scale and inter-scale approaches.

\textbf{Intra-scale methods} focus on constructing contrastive views at the identical topological granularity.
For instance, GraphCL~\cite{you2020graph} introduces four general graph augmentations (i.e., node dropping, edge perturbation, attribute masking, and subgraph sampling) to learn invariant representations by contrasting differently augmented graph views.
It systematically studies how augmentation types and strengths affect performance, revealing that appropriate compositions are crucial for effective representation learning.
However, these augmentations require tedious manual tuning and may distort graph semantics.
To alleviate this issue, SimGRACE~\cite{xia2022simgrace} eliminates the need for graph-level augmentations by perturbing the encoder parameters instead of the graph data, achieving comparable or superior performance with higher efficiency and semantic stability.

\textbf{Inter-scale methods} exploit hierarchical information through cross-granularity contrastive objectives, which can be broadly categorized into Global-Local, Global-Context, and Context-Local contrast.
DGI~\cite{velickovic2018dgi} is a pioneering work that maximizes mutual information between local node embeddings and the global graph summary, encouraging each node representation to preserve high-level structural semantics.
By contrasting real and corrupted graph pairs, DGI effectively bridges node-level and graph-level information, yielding strong performance on both transductive and inductive tasks.
Building upon this, SUBG-CON~\cite{jiao2020sub} further enhances scalability and regional awareness by contrasting central nodes with their sampled subgraphs.
Instead of relying on the complete graph, SUBG-CON focuses on regional neighborhoods, capturing mid-range dependencies while significantly reducing computational cost and enabling parallelization.

\subsection{Directed Graph Neural Network}
Research on directed graph neural networks (DiGNNs) extends traditional undirected GNNs to explicitly model edge directionality.
DGCN~\cite{tong2020dgcn} improves the message aggregation process by integrating multi-order neighborhood proximity, enabling more expressive propagation over directed structures.
DiGCN~\cite{tong2020digcn} further introduces an $\alpha$-parameterized steady-state distribution derived from personalized PageRank, formulating a principled approach for directed graph convolution.
Recent direction-aware studies have further explored magnetic operators, path-aware propagation, and scalable digraph representation learning from different perspectives~\cite{zhang2021magnet, zhou2022dhypr, geisler2023transformers_meet_digraph, li2024lightdic, su2024dirw}.

Among these advancements, MagNet~\cite{zhang2021magnet} has emerged as the dominant paradigm that leverages the magnetic Laplacian to perform convolution in the Fourier domain.
This formulation encodes the existence of edges in the magnitude of complex entries and their directional information in the phase:
\begin{equation}
\begin{aligned}
\label{eq.1}
&\;\;\;\mathbf{A}_s\left(u,v\right)=\frac{1}{2}\left(\mathbf{A}\left(u,v\right)+\mathbf{A}\left(v,u\right)\right), \\
&\mathbf{\Theta}^{\left(q\right)}\left(u,v\right):=2\pi q\left(\mathbf{A}\left(u,v\right)-\mathbf{A}\left(v,u\right)\right), \\
&\;\;\;\;\;\mathbf{L}^{\left(q\right)}:=\mathbf{D}_{s}-\mathbf{A}_{s}\odot\mathrm{exp}\left(i\mathbf{\Theta}^{\left(q\right)}\right),
\end{aligned}
\end{equation}
where $\mathbf{A}_{s}$ represents the symmetrized adjacency matrix, $\mathbf{D}_{s}$ is the corresponding degree matrix, and $\mathbf{\Theta}^{\left(q\right)}$ denotes the phase matrix controlled by the charge parameter $q\in\left(0,0.25\right)$, which determines the strength of directionality.
The real part of $\mathbf{L}^{\left(q\right)}$ indicates the edge existence, whereas the imaginary part suggests the asymmetric phases that encode edge directions.

Recently, path-based approaches have gained success in spatial-based GNNs.
DiRW~\cite{su2024dirw} introduces a weight-free, direction-aware path sampler that optimizes path generation from the perspectives of walk probability, length, and number by jointly considering node profiles and topological structure.

Despite these advances, existing DiGNNs still depend heavily on labeled data for supervision.
To alleviate this limitation, digraph contrastive learning methods have been proposed to enable unsupervised or self-supervised representation learning.
These spatial-based contrastive frameworks operate directly on digraph structure to integrate message-passing\cite{xu2018jknet, gamlp} with direction-aware augmentations.
Specifically, DiGCL\cite{tong2021directed} introduces Laplacian perturbation that perturbs the teleport probability within the approximate directed Laplacian, generating multiple contrastive views without altering the original topology.
UGCL\cite{ko2023universal} operates in the complex domain by extending the magnetic Laplacian and perturbing its phase matrix, which balances edge existence and directionality.
Signed and direction-aware variants such as SDGCL\cite{ko2023sdgcl} further show the effectiveness of Laplacian-based perturbation in more general directed settings.
Collectively, these methods demonstrate the potential of contrastive learning in capturing asymmetric structural semantics within directed graphs, yet they usually emphasize either spectral perturbation or structural view construction alone, highlighting the need for a framework that combines complementary directional views with topology-aware personalization.

\begin{figure*}[t]
\centering
\includegraphics[width=\linewidth]{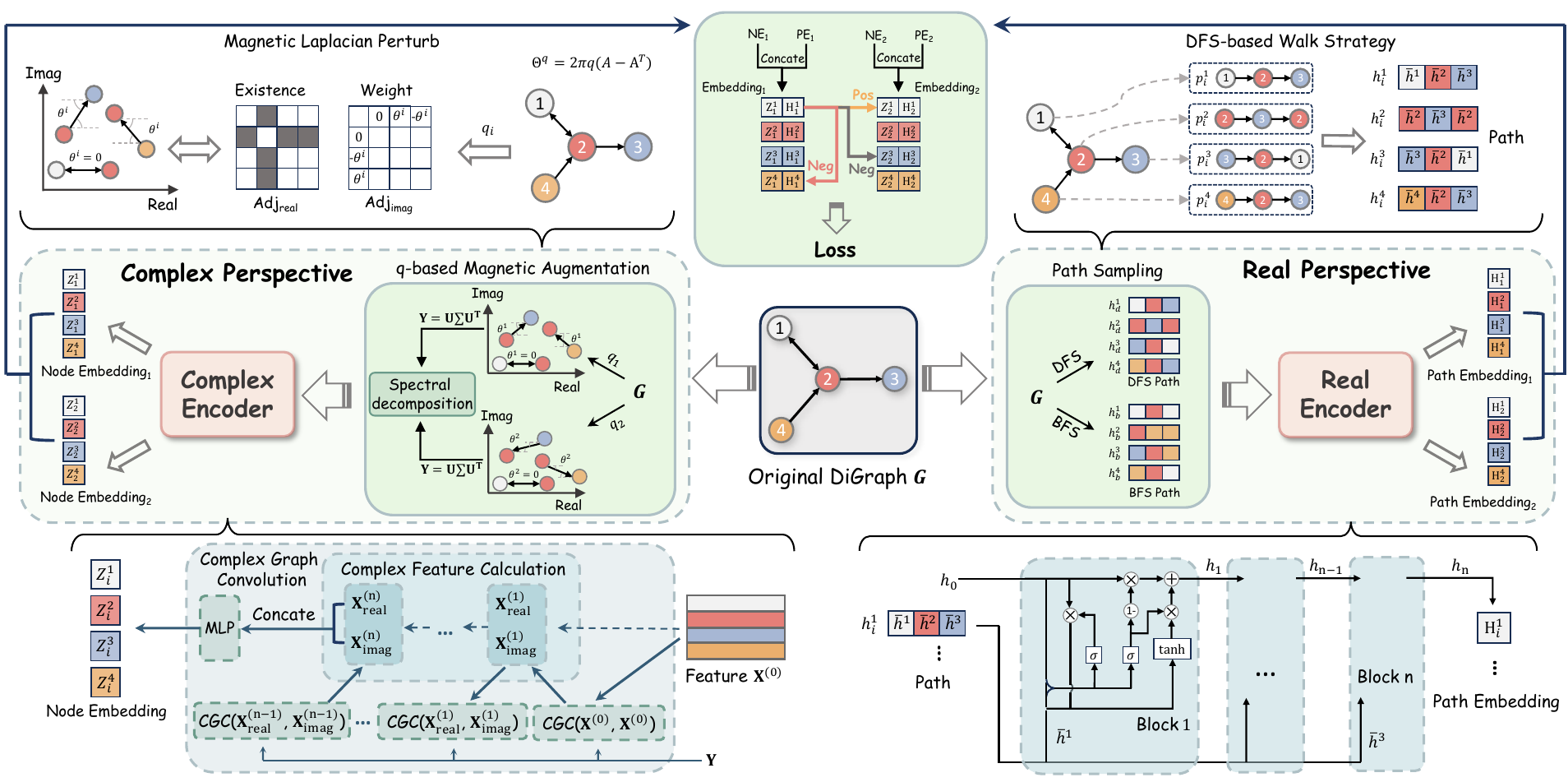}
\vspace{-0.5cm}
\caption{Overview of S2-DiGCL, including a complex-domain global personalized magnetic Laplacian augmentation (Left), a real-domain local path augmentation (Right) and a contrastive strategy that combines both perspectives (Upper).}
\vspace{-0.3cm}
\label{fig.model_framework}
\end{figure*}

\section{Proposed Method}
In this section, we present S2-DiGCL, a contrastive learning framework for digraphs that leverages dual spatial perspectives by jointly modeling perturbations in the real and complex domains.
The framework is composed of three main components:
(1) A personalized magnetic Laplacian perturbation module that generates node-level representations in the complex domain.
(2) A path-based augmentation module that captures context-level subgraph semantics in the real domain.
(3) A dual-space contrastive objective that integrates both representations to achieve comprehensive and direction-aware embedding learning.
Our design inherits the magnetic-Laplacian view construction used in prior complex-domain digraph learning and the walk-based path modeling used in spatial methods, while introducing a topology-aware personalization mechanism and a unified dual-space training pipeline.
The overall architecture of S2-DiGCL is illustrated in Fig.~\ref{fig.model_framework}.

\subsection{Complex Domain Personalized Magnetic Perturbation}
Recent developments in spectral graph theory have demonstrated that the magnetic Laplacian provides a principled formulation for representing digraphs.
Building on this foundation, several studies~\cite{furutani2020magLaplacian2, zhang2021magnet, ko2023universal} have utilized the magnetic Laplacian to encode edge directionality by introducing a complex-valued phase matrix parameterized by a charge value $q$.
In contrast to the conventional graph Laplacian, which assumes bidirectional edges, the magnetic Laplacian naturally models asymmetric dependencies through complex phases, making it particularly suitable for analyzing the structural and spectral properties of digraphs.

The main methodological extension of S2-DiGCL over prior magnetic-Laplacian-based contrastive learning lies in its \textit{personalized magnetic perturbation mechanism}, which adaptively modulates the magnetic potential field of each directed edge through a topology-aware phase coefficient.
By introducing edge-specific variations of the charge parameter $q$, this mechanism dynamically alters the phase of the magnetic Laplacian, enabling node-dependent propagation that reflects both local structural uncertainty and global directional asymmetry.
This design enriches the model’s capacity to capture diverse and complex directional patterns across different regions of the digraph.
Inspired by MAP~\cite{li2025toward}, we first measure the \textit{topological uncertainty} of each node $v$ by quantifying the balance between its in-degree and out-degree distributions:
\begin{equation}
\label{eq:uncertainty}
U_v = -\left(\frac{\tilde{d}^{\mathrm{in}}_v}{m}\log\frac{\tilde{d}^{\mathrm{in}}_v}{m}+
\frac{\tilde{d}^{\mathrm{out}}_v}{m}\log\frac{\tilde{d}^{\mathrm{out}}_v}{m}\right)\Big/ \log 2,
\end{equation}
where $m$ is the number of edges in the digraph.
A higher value of $U_v$ indicates a more balanced in-out structure, implying greater directional ambiguity.
For each directed edge $(u,v)$, we define a \textit{topological modulation coefficient} and obtain the \textit{personalized charge} as:
\begin{equation}
\label{eq:q_topo}
q^{\star}_{uv}=q_0 \cdot q^{\mathrm{topo}}_{uv},\;\;\;
q^{\mathrm{topo}}_{uv}=\tanh\left(\frac{U_u + U_v}{\operatorname{mean}(U)}\right),
\end{equation}
which emphasizes edges connecting nodes with high directional uncertainty.
Here $q_0 \in (0,0.25]$ is a global base charge controlling the overall phase magnitude.
This formulation provides an edge-specific adaptation of the magnetic potential, ensuring that structurally ambiguous regions receive stronger directional perturbations.

To incorporate stochastic directional variability, we further introduce a \textit{probabilistic perturbation factor} defined as:
\begin{equation}
\label{eq:phi}
\mathbf{\Phi}^{(q^{\star})}_r(u,v) =
\begin{cases}
q^{\star}_{uv}, & \text{with probability } r, \\
1-q^{\star}_{uv}, & \text{with probability } 1-r,
\end{cases}
\end{equation}
where $r \in [0,1]$ controls the likelihood of maintaining the original edge orientation.
A larger $r$ favors the original directionality, while a smaller $r$ introduces greater structural perturbation.
This stochastic mechanism implicitly reverses edge directions in a probabilistic manner, introducing diversity without explicit graph rewiring.
The resulting phase matrix integrates both directionality and uncertainty:
\begin{equation}
\label{eq:theta}
\mathbf\Theta^{(q^{\star})}_r(u,v)=
2\pi \left(\mathbf A(u,v) - \mathbf A(v,u)\right)
\odot \mathbf\Phi^{\left(q^{\star}\right)}_r\left(u,v\right),
\end{equation}
where $\odot$ denotes the element-wise product.
Notably, it is easy to see that $\mathbf{\Theta}^{(q^{\star})}(u,v)=-\mathbf{\Theta}^{(1-q^{\star})}(v,u)$ from Eq.~(\ref{eq.1}), meaning that simply perturbing $q^{\star}$ is equivalent to topologically reversing edge directions in a probabilistic manner.
The random perturbation in $\boldsymbol{\Phi}^{(q^{\star})}_r$ actually corresponds to a random perturbation of the direction of the directed edges, reflecting the concept of personalized perturbation without explicit edge redirecting.
Beyond topological perturbations, we further introduce a Laplacian augmentation, where $q^{\star}$ is perturbed by a small term $\Delta q$.
This yields the \textit{\textbf{personalized magnetic Laplacian perturbation}}:
\begin{equation} 
\label{eq.4} 
\mathbf{L}^{(q^{\star})}_{r,\Delta q} := \mathbf{D}_{s} - \mathbf{A}_{s} \odot \mathrm{exp}\left(i\mathbf{\Theta}^{(q^{\star}+\Delta q)}_{r}\right),
\end{equation}
where $\Delta q$ is a perturbation term that satisfies $q^{\star}+\Delta q\in [0, 0.25]$ to avoid excessive phase shifts.
Through this operation, the sparsity of the magnetic Laplacian matrix is maintained, which means it can be directly used to the subsequent GNN-based encoders.

With the personalized magnetic Laplacian in place, we follow the standard GCL protocol based on paired spectral views.
Specifically, S2-DiGCL constructs two correlated Laplacian views, the original $\mathbf{L}^{(q^{\star})}$ and the perturbed $\mathbf{L}^{(q^{\star})}_{r,\Delta q}$.
These views are fed into a shared complex-valued graph convolutional encoder, where each layer updates both the real and imaginary parts of node features.

Taking $\mathbf{L}^{(q^{\star})}$ as an example, the $l$-th layer is written as:
\begin{equation}
\label{eq.5}
\begin{aligned}
&\;\mathbf{X}^{(l)}_{\mathrm{real}} = \left(\mathbf{D}_{s}^{-\frac{1}{2}} \mathbf{A}_{s}\mathbf{D}_{s}^{-\frac{1}{2}} \odot \exp\left(i\mathbf{\Theta}^{(q^{\star})}\right)\right) \mathbf{X}^{(l-1)}_{\mathrm{real}}, \\
&\mathbf{X}^{(l)}_{\mathrm{imag}} = \left(\mathbf{D}_{s}^{-\frac{1}{2}} \mathbf{A}_{s}\mathbf{D}_{s}^{-\frac{1}{2}} \odot \exp \left(i\mathbf{\Theta}^{(q^{\star})}\right)\right) \mathbf{X}^{(l-1)}_{\mathrm{imag}},
\end{aligned}
\end{equation}
with $\mathbf{X}^{(0)}_{\mathrm{real}}=\mathbf{X}^{(0)}_{\mathrm{imag}}=\mathbf{X}^{(0)}$ denoting the initial node features.

After stacking $L$ convolutional layers, we flatten the complex output by concatenating the two parts and obtain the final node embeddings via an MLP:
\begin{equation}
\label{eq.6}
\mathrm{NE}=\mathrm{MLP}\left(\mathbf{X}^{(L)}_{\mathrm{real}}\Vert\mathbf{X}^{(L)}_{\mathrm{imag}}\right),
\end{equation}
where $\Vert$ denotes concatenation.
The same encoder is applied to the perturbed view $\mathbf{L}^{(q^{\star})}_{r,\Delta q}$, yielding a correlated pair of complex-domain representations for contrastive learning.

\subsection{Real Domain Path Augmentation}
While the complex-domain magnetic Laplacian perturbation captures global directional patterns through phase shifts, the local flow of information within digraphs is equally important for tasks that require fine-grained structural awareness.
Recent work~\cite{su2024dirw} has shown that walk-based approaches are particularly effective at encoding such local directionality by explicitly modeling node sequences along directed paths.

To complement the global complex-domain representation, we propose a path augmentation framework that generates subgraph-level views in the real domain.
This component is inspired by second-order biased walks such as Node2Vec~\cite{Node2vec} and direction-aware path modeling in recent digraph learning methods~\cite{su2024dirw}, but is incorporated here as a local view generator for contrastive learning.
For each node, we employ both Breadth-First Search (BFS) and Depth-First Search (DFS) strategies to sample two distinct directed walk sequences.
Guided by the homophily principle~\cite{mcpherson200homophily_theory1}, BFS-generated sequences capture local structure patterns that are homophilous with the central node, whereas DFS-generated sequences explore longer-range heterophilous neighbors.

To control the path sampling behavior, we introduce a second-order direction-aware sampler characterized by two parameters $p$ and $q$, which govern the trade-off between local and global exploration.
Consider a random walk that has just traversed the edge $(u,v)$ and currently resides at node $v$.
For all outgoing edges $(v,x)$, the unnormalized transition probability $\alpha$ is defined as:
\begin{equation}
\alpha(v, x) =
\begin{cases}
1/p, & \text{if } d_{ux}=0, \\
1,   & \text{if } d_{ux}=1, \\
1/q, & \text{if } d_{ux}=2, \\
0,   & \text{otherwise},
\end{cases}
\end{equation}
where $d_{ux}$ denotes the shortest-path distance between $u$ and $x$.
Setting $0<p<1$ and $q>1$ biases the walk toward BFS-like local exploration, whereas $p>1$ and $0<q<1$ promote DFS-like deeper traversal.
For each node $v_i$ in the digraph, we generate two sequences $h_b^i$ and $h_d^i$ based on BFS and DFS:
\begin{equation}
\begin{aligned}
&\mathbf{P}_{b} = \left\{ h_b^1, h_b^2,\dots, h_b^n \right\}, \\
&\mathbf{P}_{d} = \left\{ h_d^1, h_d^2,\dots, h_d^n \right\},
\end{aligned}
\end{equation}
where $h_b^i$ denotes the BFS-based path sequence sampled with node $v_i$ as the central node.
Similarly, $h_d^i$ represents the DFS-based path sequence derived from the node $v_i$.

Given a path list $\mathbf{P}$, the path aggregator $f_{path}$ transforms the sampled path sequences into path embeddings (PE).
These embeddings are designed to effectively capture the intricate local topology surrounding the central node, thereby providing a rich representation of the node’s neighborhood.
Crucially, the aggregator must preserve both the node order and directionality of the sequences to maintain structural fidelity.
We implement $f_{path}$ as a gated recurrent unit (GRU)~\cite{cho2014learning}, a standard sequential encoder that is particularly suitable for preserving ordered directional dependencies within each sampled path.
It processes each sequence stepwise, updating a hidden state that encapsulates the path’s directional context up to the current node.
For a sequence $h^i=\left\{v^i_1,v^i_2,\dots,v^i_l\right\}$, the GRU computes:
\begin{equation}
\begin{aligned}
&\;\;\;\;\;\; \mathbf{z}^t = \sigma\left(\mathbf{W}_{z} \left[\mathbf{h}^{t-1}, \mathbf{x}^t\right]\right), \\
&\;\;\;\;\;\; \mathbf{r}^t = \sigma\left(\mathbf{W}_{r} \left[\mathbf{h}^{t-1}, \mathbf{x}^t\right]\right), \\
& \hat{\mathbf{h}}^t = \tanh\left(\mathbf{W}_{h}\left[\mathbf{r}^t\odot\mathbf{h}^{t-1}, \mathbf{x}^t\right]\right), \\
&\;\; \mathbf{h}^t=\left(1-z^t\right)\odot\mathbf{h}^{t-1}+z^t\odot\hat{\mathbf{h}}^t,
\end{aligned}
\end{equation}
where $\mathbf{z}^t$ and $\mathbf{r}^t$ are update and reset gates, respectively, and $\mathbf{h}^t$ is the hidden state at step $t$.
The final embedding for the sequence is derived from the last hidden state $\mathbf{h}^l$, which encodes the cumulative directional flow of the entire path.

\subsection{Contrastive Objective}
To unify the global and local representations, we fuse the embeddings obtained from the two spatial perspectives.
Specifically, the node embeddings $\mathrm{NE}_1$ and $\mathrm{NE}_2$ are produced from the complex-domain personalized magnetic Laplacian perturbation, while the path embeddings $\mathrm{PE}_1$ and $\mathrm{PE}_2$ are generated from the real-domain path augmentation.
The final representations are formed by concatenating the two types of embeddings and passing them through a projection head:
\begin{equation}
\begin{aligned}
\mathrm{E}_1 &= g\left(\mathrm{NE}_1\Vert\mathrm{PE}_1\right),\\
\mathrm{E}_2 &= g\left(\mathrm{NE}_2\Vert\mathrm{PE}_2\right),
\end{aligned}
\end{equation}
where $g(\cdot)$ denotes a nonlinear projection that maps the fused embeddings into a shared latent space suitable for contrastive learning.
We adopt this lightweight fusion strategy to preserve the complementary information from the two branches while avoiding the additional parameters and optimization instability introduced by more elaborate cross-view fusion modules.
The goal of the contrastive objective is to maximize agreement between two correlated views of the same node while distinguishing them from other nodes.
Let $\mathbf{h}_i^1$ and $\mathbf{h}_i^2$ denote the embeddings of the $i$-th node in $\mathrm{E}_1$ and $\mathrm{E}_2$, respectively.
The pair $\left(\mathbf{h}_i^1,\mathbf{h}_i^2\right)$ is treated as a positive sample, and all other combinations are treated as negatives.
As in many InfoNCE-based graph contrastive methods, this formulation may still contain semantically related nodes among negatives; we therefore view the current objective as a simple and effective baseline choice, and leave more refined hard-negative or false-negative-aware strategies to future work.
In this way, the model is encouraged to produce view-invariant representations that capture consistent semantics across both spatial domains.
To quantify this alignment, we employ the InfoNCE~\cite{oord2018representation} objective, which estimates a lower bound of the mutual information (MI) between the two embedding distributions.
\begin{table*}[t]
\centering
\caption{Statistics of the datasets used in experiments.}
\vspace{-2mm}
\label{tab:dataset_stats}
\resizebox{0.85\linewidth}{!}{
\begin{tabular}{c|ccccccc}
\toprule
Dataset & \#Nodes & \#Edges & \#Features & \#Classes & Train/Valid/Test & Description \\
\midrule
CoraML & 2,995 & 8,416 & 2,879 & 7 & 140/500/2355 & Citation network \\
Citeseer & 3,312 & 4,591 & 3,703 & 6 & 120/500/2692 & Citation network \\
Pubmed & 19,717 & 88,648 & 500 & 3 & 60/500/19157 & Citation network \\
Actor & 7,600 & 26,659 & 932 & 5 & 3648/2432/1520 & Co-occurrence network \\
WikiCS & 11,701 & 290,519 & 300 & 10 & 580/1769/5847 & Weblink network \\
Chameleon & 890 & 13,534 & 2,325 & 5 & 409/287/194 & Weblink network \\
Squirrel & 2,223 & 65,578 & 2,089 & 5 & 1053/718/452 & Weblink network \\
\bottomrule
\end{tabular}}
\vspace{-2mm}
\end{table*}
The total loss consists of two components:
\begin{equation}
\begin{aligned}
&\;\;\;\mathcal{L}_{\mathrm{inter}} = -\frac{1}{n} \sum_{i=1}^{n} 
\log \frac {\exp\!\left(\mathcal{S}\!\left(\mathbf{h}_i^1,\mathbf{h}_i^2\right)/\tau\right)}
{\sum_{j=1}^{n}\exp\!\left(\mathcal{S}\!\left(\mathbf{h}_i^1,\mathbf{h}_j^2\right)/\tau\right)},\\
&\mathcal{L}_{\mathrm{intra}} = -\frac{1}{2n} \sum_{k=1}^{2} \sum_{i=1}^{n} 
\log \frac {1}{\sum_{j\neq i} \exp\!\left(\mathcal{S}\!\left(\mathbf{h}_i^k,\mathbf{h}_j^k\right)/\tau\right)},\\
&\quad\quad\;\;\;\;\;\;\;\;\;\;\;\;\;\quad \mathcal{L} = \mathcal{L}_{\mathrm{inter}} + \mathcal{L}_{\mathrm{intra}},
\end{aligned}
\end{equation}
where $\mathcal{S}(\cdot,\cdot)$ denotes cosine similarity and $\tau$ is a temperature coefficient controlling the concentration of the similarity distribution.

Here, $\mathcal{L}_{\mathrm{inter}}$ enforces consistency between the complex-domain and real-domain views by pulling positive pairs closer across modalities, while $\mathcal{L}_{\mathrm{intra}}$ maintains local smoothness within each view by discouraging over-collapsed representations.
Jointly optimizing these two terms enables S2-DiGCL to balance semantic alignment and structural diversity, ensuring that the learned embeddings are simultaneously discriminative and direction-aware across both spatial domains.

\section{Experiments}
In this section, we present a comprehensive empirical evaluation of S2-DiGCL to validate its effectiveness across multiple downstream tasks and diverse real-world digraph datasets.
Our experiments are systematically designed to address the following four key research questions:
\textbf{Q1:}
Does S2-DiGCL achieve superior performance compared to existing unsupervised and supervised baselines?
\textbf{Q2:}
How do the real-domain path augmentation and complex-domain magnetic perturbation modules individually and jointly contribute to the model’s performance?
\textbf{Q3:}
How robust is S2-DiGCL to variations in key hyperparameters?
\textbf{Q4:}
How efficient is S2-DiGCL in terms of training time and convergence speed?

\subsection{Experimental Setup}
\label{sec:Experimental Setup}

\subsubsection{Datasets}
To comprehensively evaluate the proposed framework, we conduct experiments on seven widely used real-world directed graph datasets, covering three distinct network domains.
These datasets collectively cover a broad spectrum of structural characteristics, including varying graph scales, feature dimensionalities, and degrees of homophily and heterophily.
Table~\ref{tab:dataset_stats} summarizes their key statistics.

\subsubsection{Baselines}
We compare S2-DiGCL with a broad range of baseline methods covering both supervised and unsupervised paradigms, as well as undirected and directed graph settings.
To ensure a fair and comprehensive evaluation, we group these baselines into four categories according to their learning paradigm and graph type.
\textit{(i) Supervised undirected GNNs} include
GCN~\cite{kipf2016gcn}, GAT~\cite{velivckovic2017gat}, and 
RAW-GNN~\cite{ijcai22_rawgnn}.
\textit{(ii) Supervised directed GNNs} include
DiGCN~\cite{tong2020digcn}, and MagNet~\cite{zhang2021magnet}.
\textit{(iii) Unsupervised undirected GCLs} include
DGI~\cite{velickovic2018dgi}, GCA~\cite{zhu2021graph}, GraphCL~\cite{hafidi2022negative}, and GRACE~\cite{zhu2020deep}.
\textit{(iv) Unsupervised directed GCLs} include
DiGCL~\cite{tong2021directed}, and UGCL~\cite{ko2023universal}.

\subsubsection{Implementation Details}
For link prediction, we follow prior work\cite{zhang2021magnet, li2024lightdic} with 80\%/15\%/5\% train/validation/test edge splits.
For unsupervised methods, labels are used only for linear evaluation\cite{velickovic2018dgi}.
After unsupervised pretraining, encoders are frozen and logistic regression is trained on embeddings.
To ensure reproducibility, the complete source code for implementing the S2-DiGCL is avaliable at \url{https://anonymous.4open.science/r/S2-DiGCL-169D}.

\subsubsection{Experimental Environment}
All experiments are conducted on an Intel(R) Xeon(R) Platinum 8468V CPU and an NVIDIA H800 PCIe GPU, running CUDA 12.2.
The operating system is Ubuntu 20.04.6 LTS, and the software stack consists of Python 3.8 and PyTorch 2.2.1.
All baseline implementations are based on their official repositories or open-source reimplementations, and hyperparameters are tuned to ensure fairness.

\begin{table*}[t]
\centering
\caption{Node classification performance. The best result is \colorbox{blue!15!white}{\textbf{bold}}. The second result is \underline{underlined}.}
\vspace{-3mm}
\label{tab:node_clf}
\resizebox{\linewidth}{!}{
\begin{tabular}{cc|ccccccc}
\toprule
\multicolumn{2}{c}{Method} & CoraML & Citeseer & WikiCS & Actor & Pubmed & Chameleon & Squirrel \\
\midrule
\multirow{5}*{\rotatebox{90}{SUPERVISED}} 
& GCN & \underline{80.90$\pm$0.36} & \underline{65.57$\pm$0.60} & 77.56$\pm$0.29 & 31.20$\pm$0.51 & 78.19$\pm$0.17 & 43.20$\pm$1.80 & 37.74$\pm$1.08 \\
& GAT & 80.17$\pm$1.09 & 63.76$\pm$1.71 & 77.00$\pm$0.83 & 29.41$\pm$1.25 & \colorbox{blue!15!white}{\textbf{79.83$\pm$0.30}} & 41.44$\pm$4.04 & 34.20$\pm$1.24 \\
& RAW-GNN & 76.51$\pm$0.87 & 59.38$\pm$0.81 & 75.09$\pm$0.26 & \underline{35.08$\pm$0.53} & 76.45$\pm$0.31 & 42.78$\pm$0.96 & 38.45$\pm$2.92 \\
\cmidrule(lr){2-9}
& DiGCN & 77.94$\pm$0.35 & 62.24$\pm$0.50 & 77.63$\pm$0.46 & 33.91$\pm$1.02 & 76.60$\pm$0.20 & 40.52$\pm$3.43 & 37.65$\pm$1.14 \\
& MagNet & 79.41$\pm$1.05 & 63.91$\pm$1.27 & 76.74$\pm$0.70 & 31.26$\pm$0.67 & 77.34$\pm$0.63 & 43.09$\pm$2.77 & 40.97$\pm$1.11 \\
\midrule
\multirow{8}*{\rotatebox{90}{UNSUPERVISED}} 
& DGI & 72.55$\pm$2.29 & 57.30$\pm$4.07 & 74.19$\pm$2.90 & 30.67$\pm$0.63 & 74.32$\pm$1.87 & 39.95$\pm$1.76 & 34.73$\pm$1.24 \\
& GCA & 80.39$\pm$1.11 & 65.30$\pm$1.38 & 72.82$\pm$1.90 & 29.39$\pm$0.97 & 71.05$\pm$4.04 & 39.38$\pm$0.86 & \underline{41.72$\pm$1.13} \\
& GraphCL & 74.93$\pm$1.68 & 62.65$\pm$1.71 & \underline{78.37$\pm$0.77} & 31.51$\pm$1.54 & 73.20$\pm$1.82 & 43.59$\pm$1.38 & 39.29$\pm$3.60 \\
& GRACE & 78.34$\pm$0.84 & 63.49$\pm$1.33 & 70.49$\pm$3.13 & 30.72$\pm$0.63 & 74.15$\pm$2.42 & 41.84$\pm$2.08 & 40.75$\pm$1.01 \\
\cmidrule(lr){2-9}
& UGCL & 78.66$\pm$1.52 & 64.20$\pm$1.70 & 75.43$\pm$3.15 & 33.63$\pm$0.70 & 76.15$\pm$4.02 & \underline{44.49$\pm$1.43} & 39.99$\pm$1.57 \\
& DiGCL & 77.83$\pm$2.57 & 63.36$\pm$2.23 & 74.83$\pm$2.88 & 34.53$\pm$0.42 & 79.14$\pm$1.16 & 42.66$\pm$1.21 & 38.50$\pm$0.68 \\
\cmidrule(lr){2-9}
& \textbf{S2-DiGCL} & \colorbox{blue!15!white}{\textbf{81.83$\pm$1.37}} & \colorbox{blue!15!white}{\textbf{67.68$\pm$0.63}} & \colorbox{blue!15!white}{\textbf{78.42$\pm$1.13}} & \colorbox{blue!15!white}{\textbf{35.71$\pm$0.84}} & \underline{79.16$\pm$0.81} & \colorbox{blue!15!white}{\textbf{45.18$\pm$2.42}} & \colorbox{blue!15!white}{\textbf{41.95$\pm$1.46}} \\
\bottomrule
\end{tabular}}
\end{table*}

\begin{table*}[t]
\centering
\caption{Link prediction performance. The best result is \colorbox{blue!15!white}{\textbf{bold}}. The second result is \underline{underlined}.}
\vspace{-3mm}
\label{tab:link_pred}
\resizebox{\linewidth}{!}{
\begin{tabular}{cccccccccc}
\toprule
& & \multicolumn{4}{c}{Existence} & \multicolumn{4}{c}{Direction} \\
\cmidrule(r){3-6} \cmidrule(l){7-10}
& & CoraML & Citeseer & Chameleon & Squirrel & CoraML & Citeseer & Chameleon & Squirrel \\
\midrule
\multirow{5}*{\rotatebox{90}{SUPERVISED}} & GCN & 75.81$\pm$0.58 & \underline{69.87$\pm$0.60} & 84.05$\pm$0.42 & 87.24$\pm$0.10 & 84.01$\pm$0.26 & 82.46$\pm$2.12 & 87.50$\pm$0.53 & 87.22$\pm$0.17 \\
& GAT & 72.27$\pm$1.50 & 67.16$\pm$1.55 & 83.17$\pm$1.89 & 85.88$\pm$1.96 & 82.56$\pm$2.53 & 79.91$\pm$2.09 & 87.18$\pm$2.70 & 86.98$\pm$1.96 \\
& RAW-GNN & 77.69$\pm$0.60 & 69.04$\pm$0.96 & \underline{84.47$\pm$0.58} & 88.43$\pm$0.16 & \underline{89.72$\pm$1.21} & \colorbox{blue!15!white}{\textbf{86.79$\pm$1.25}} & \underline{90.43$\pm$1.20} & 88.95$\pm$0.15 \\
\cmidrule(lr){2-10}
& DiGCN & 78.90$\pm$0.16 & 67.21$\pm$0.36 & 83.82$\pm$0.39 & 88.28$\pm$0.14 & 88.20$\pm$0.38 & 86.29$\pm$1.52 & 90.21$\pm$0.69 & 84.91$\pm$0.10 \\
& MagNet & \underline{79.38$\pm$0.67} & 69.48$\pm$1.19 & 83.73$\pm$0.25 & \underline{88.47$\pm$0.12} & 89.57$\pm$0.42 & 86.07$\pm$1.43 & 90.32$\pm$0.58 & 86.90$\pm$0.28 \\
\midrule
\multirow{8}*{\rotatebox{90}{UNSUPERVISED}} & DGI & 72.12$\pm$0.98 & 63.65$\pm$1.98 & 80.30$\pm$1.02 & 84.66$\pm$2.97 & 85.71$\pm$1.36 & 77.59$\pm$1.51 & 84.10$\pm$4.03 & 83.77$\pm$2.57 \\
& GCA & 75.93$\pm$1.06 & 64.32$\pm$0.77 & 82.69$\pm$0.64 & 86.63$\pm$0.26 & 80.87$\pm$0.94 & 76.79$\pm$1.64 & 82.34$\pm$1.20 & 83.45$\pm$0.32 \\
& GraphCL & 75.74$\pm$0.89 & 66.29$\pm$2.26 & 82.12$\pm$0.81 & 83.06$\pm$1.81 & 88.37$\pm$0.54 & 77.05$\pm$2.91 & 89.47$\pm$1.04 & \underline{89.17$\pm$0.31} \\
& GRACE & 74.31$\pm$0.63 & 65.76$\pm$1.12 & 81.91$\pm$0.82 & 86.57$\pm$0.21 & 80.03$\pm$0.74 & 85.68$\pm$2.56 & 83.24$\pm$2.29 & 83.01$\pm$0.77 \\
\cmidrule(lr){2-10}
& UGCL & 73.64$\pm$1.09 & 66.55$\pm$2.01 & 79.47$\pm$0.98 & 83.93$\pm$1.36 & 85.52$\pm$1.68 & 79.51$\pm$1.04 & 87.44$\pm$1.87 & 84.60$\pm$4.91 \\
& DiGCL & 76.48$\pm$1.59 & 64.72$\pm$2.32 & 81.66$\pm$0.94 & 85.23$\pm$0.24 & 83.32$\pm$3.39 & 76.88$\pm$1.61 & 86.65$\pm$0.85 & 82.48$\pm$1.11 \\ 
\cmidrule(lr){2-10}
& \textbf{S2-DiGCL} & \colorbox{blue!15!white}{\textbf{79.60$\pm$1.45}} & \colorbox{blue!15!white}{\textbf{71.57$\pm$1.25}} & \colorbox{blue!15!white}{\textbf{87.57$\pm$0.89}} & \colorbox{blue!15!white}{\textbf{89.46$\pm$0.40}} & \colorbox{blue!15!white}{\textbf{89.80$\pm$0.75}} & \underline{86.37$\pm$1.69} & \colorbox{blue!15!white}{\textbf{91.04$\pm$2.13}} & \colorbox{blue!15!white}{\textbf{89.76$\pm$1.54}} \\
\bottomrule
\end{tabular}}
\vspace{-2mm}
\end{table*}

\subsection{Overall Performance}
To address \textbf{Q1}, we conduct extensive experiments comparing S2-DiGCL with all baseline methods across three representative tasks: node classification, link prediction, and t-SNE-based cluster visualization.
These tasks jointly evaluate the discriminative power, structural awareness, and embedding interpretability of the learned node representations.

\subsubsection{Node Classification Performance}
To assess the quality and generalizability of the learned node embeddings, we perform node classification experiments on all benchmark datasets.
The results are summarized in Table~\ref{tab:node_clf}.

We observe that conventional unsupervised baselines (e.g., DGI, GRACE), which are primarily designed for undirected graphs, exhibit substantial performance degradation on directed datasets.
This performance drop can be attributed to their inherent assumption of bidirectional message passing and their inability to explicitly model asymmetric edge relationships.
In contrast, direction-aware contrastive methods such as DiGCL and UGCL achieve notable improvements by incorporating spectral or complex-domain perturbations that respect edge directionality.
However, their augmentation strategies are limited to a single representation space, either real or complex, which constrains their ability to capture multi-domain relational semantics.
Our proposed S2-DiGCL bridges this gap by jointly modeling real-domain path augmentations and complex-domain magnetic perturbations, effectively integrating structural and directional cues in a complementary manner.
This dual-domain learning scheme yields the best performance on six out of seven datasets.
Despite being entirely unsupervised, S2-DiGCL attains comparable or even superior performance to several supervised GNNs, including RAW-GNN and MagNet, demonstrating that it can capture task-relevant structure without relying on labeled data.
Across five independent runs, the standard deviations remain moderate on most datasets, indicating that the learned representations are reasonably stable under different random seeds.

\subsubsection{Link Prediction Performance}
To further examine the generality of our framework, we conduct link prediction experiments on four representative datasets: CoraML, Citeseer, Chameleon, and Squirrel.
This task evaluates whether the learned representations can effectively capture both topological connectivity and directional dependencies between nodes.
We adopt two evaluation settings: \textit{Existence}, which predicts whether an edge exists between two nodes, and \textit{Direction}, which determines the orientation of an existing edge.

As summarized in Table~\ref{tab:link_pred}, our proposed S2-DiGCL consistently outperforms both supervised and unsupervised baselines in the Existence setting, demonstrating the efficacy of the dual-spatial augmentation strategy in learning coherent node embeddings that generalize across missing or perturbed connections.
In the more challenging Direction setting, unsupervised methods tailored for digraphs (e.g., UGCL, DiGCL) exhibit superior performance over their undirected counterparts (e.g., DGI, GCA, GraphCL, GRACE), confirming the importance of explicitly modeling edge asymmetry in digraphs.
Building upon these approaches, S2-DiGCL further enhances directional prediction accuracy by jointly integrating real-domain path augmentations with complex-domain magnetic perturbations.
This dual-domain formulation allows the model to simultaneously capture global connectivity cues and fine-grained directional dependencies.
These results highlight that S2-DiGCL learns discriminative node embeddings without label supervision and transfers effectively to both link existence and link direction prediction on the four evaluated datasets.
\begin{figure*}
\centering
\subfigure[DGI]{\includegraphics[width=0.19\textwidth]{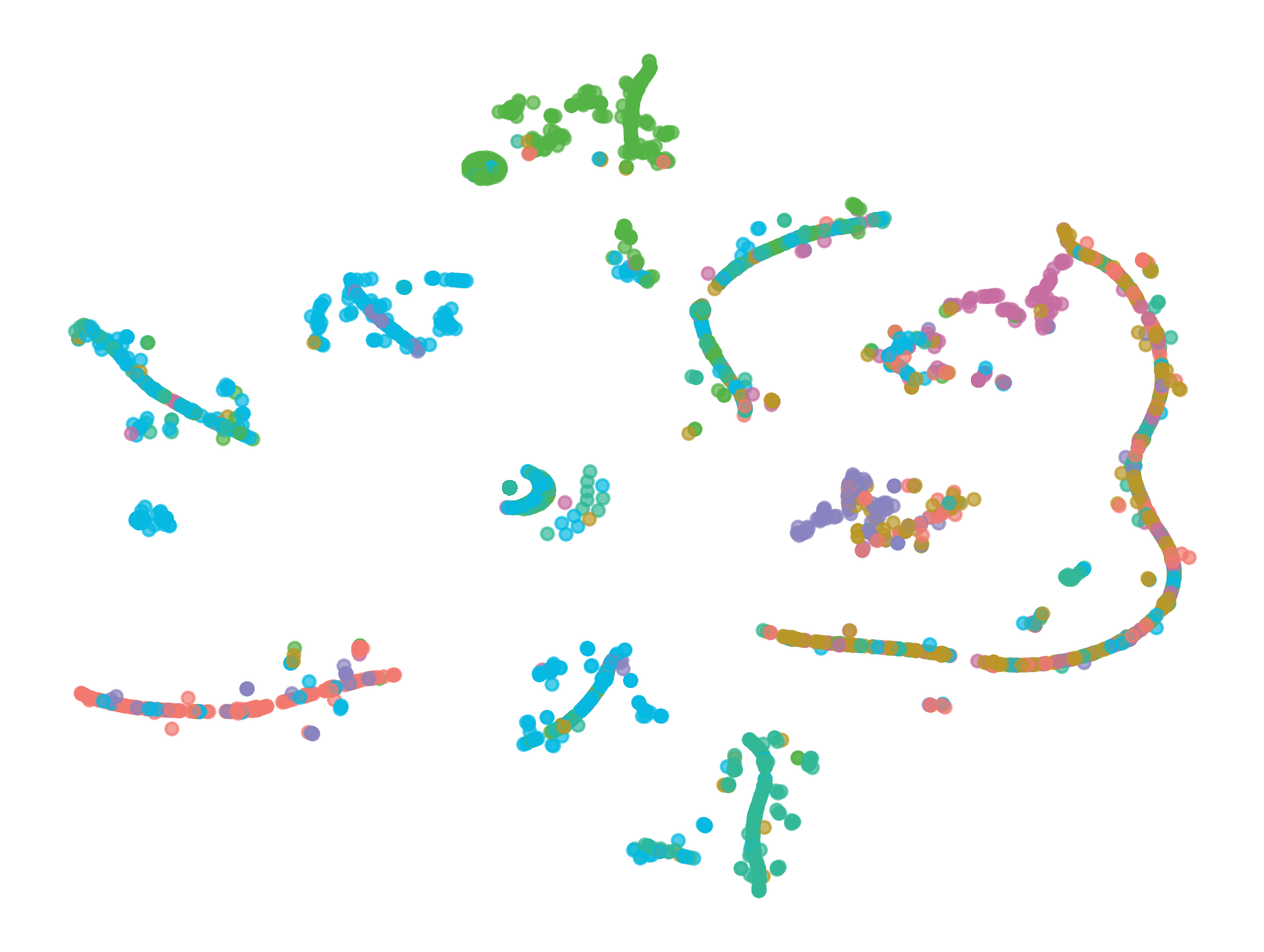}}
\subfigure[GraphCL]{\includegraphics[width=0.19\textwidth]{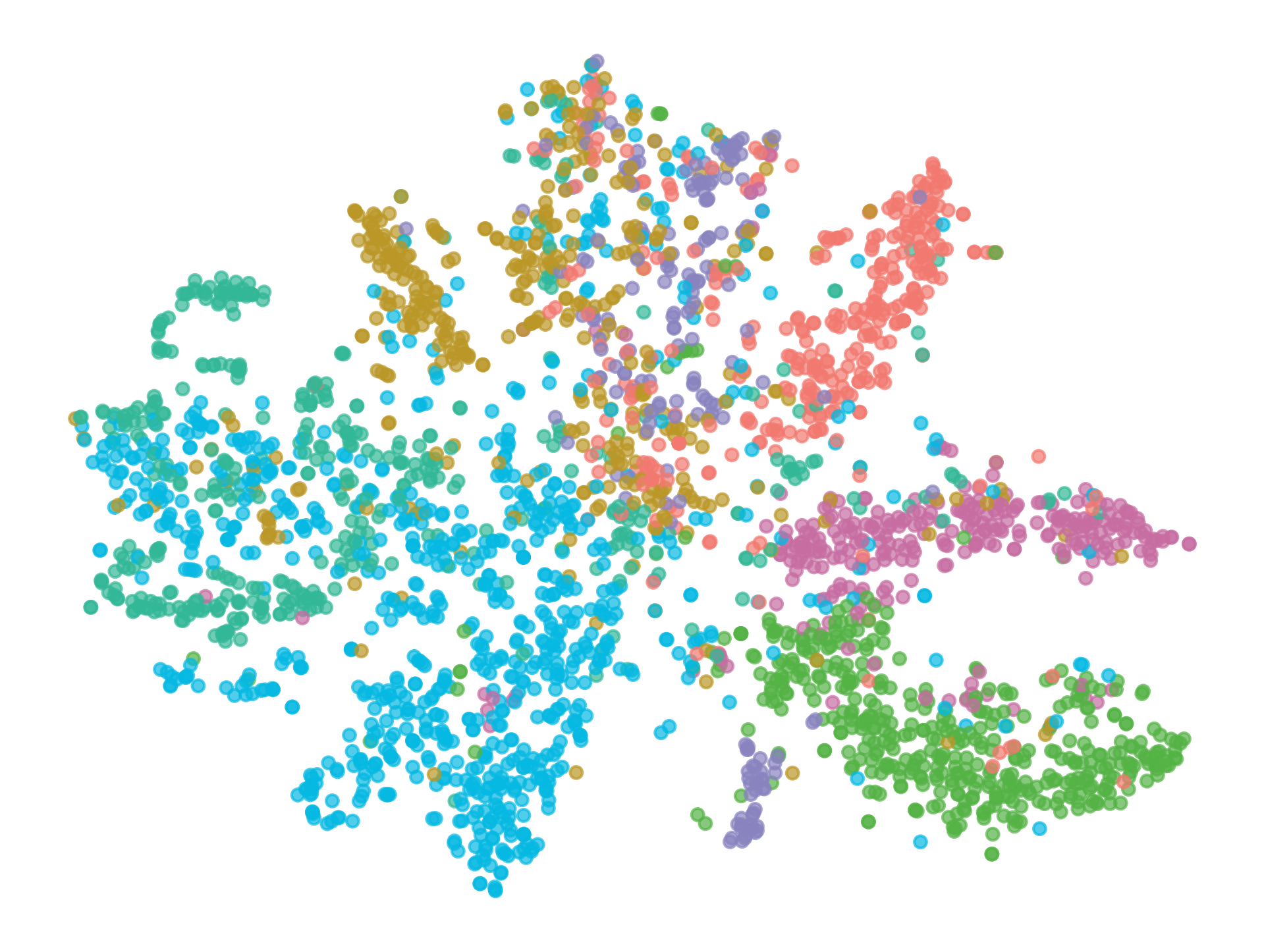}}
\subfigure[UGCL]{\includegraphics[width=0.19\textwidth]{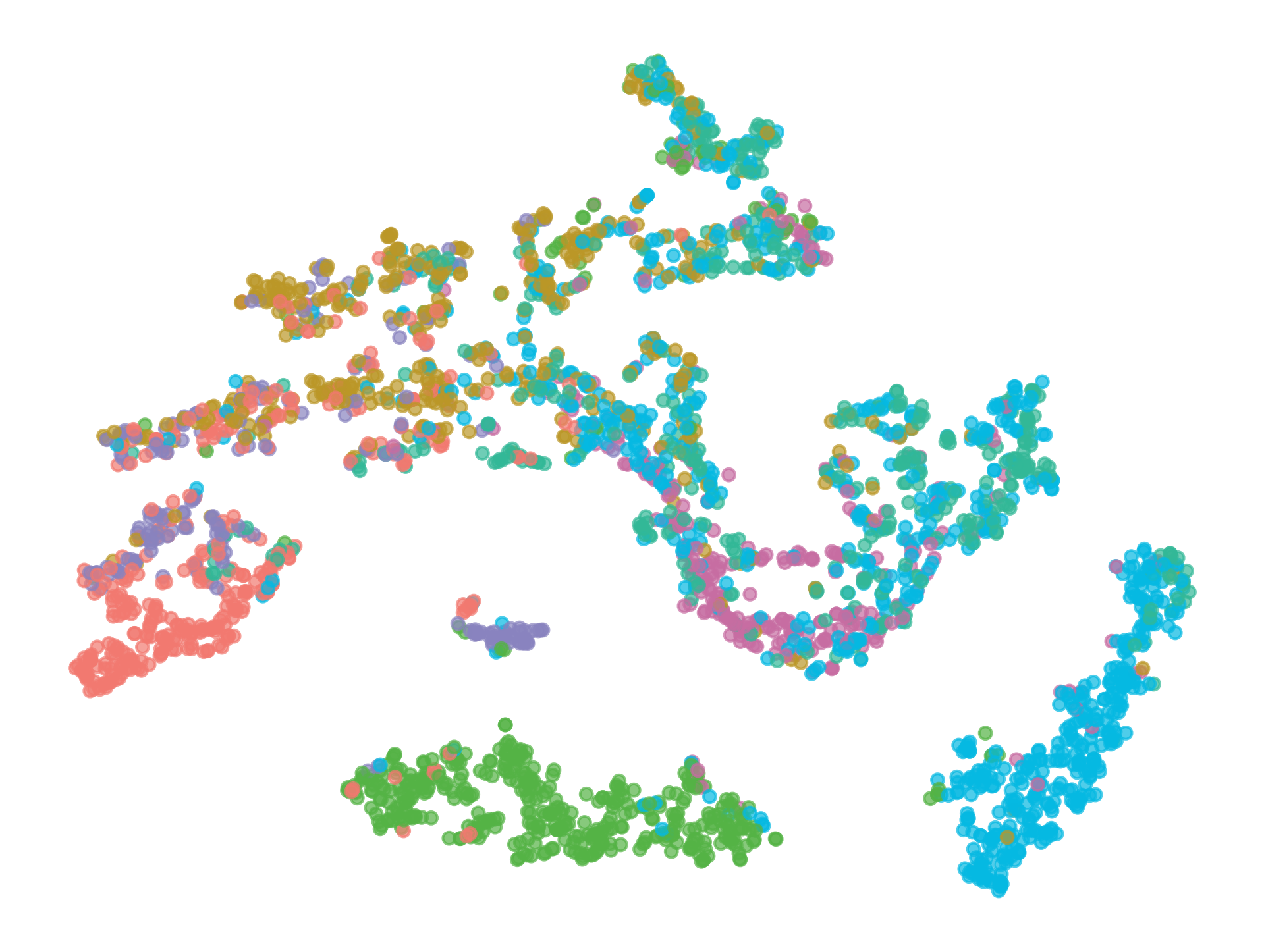}}
\subfigure[DiGCL]{\includegraphics[width=0.19\textwidth]{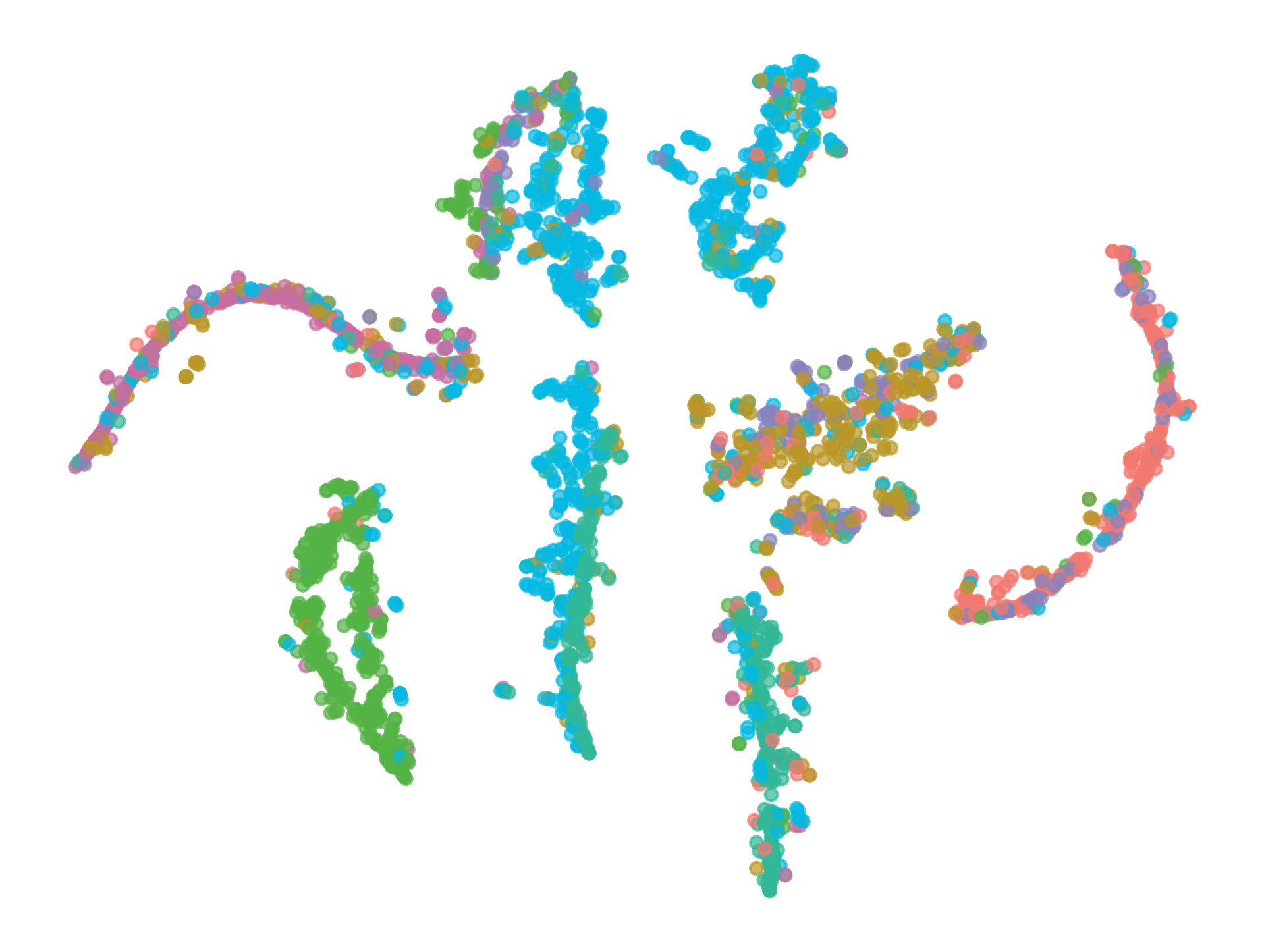}}
\subfigure[S2-DiGCL]{\includegraphics[width=0.19\textwidth]{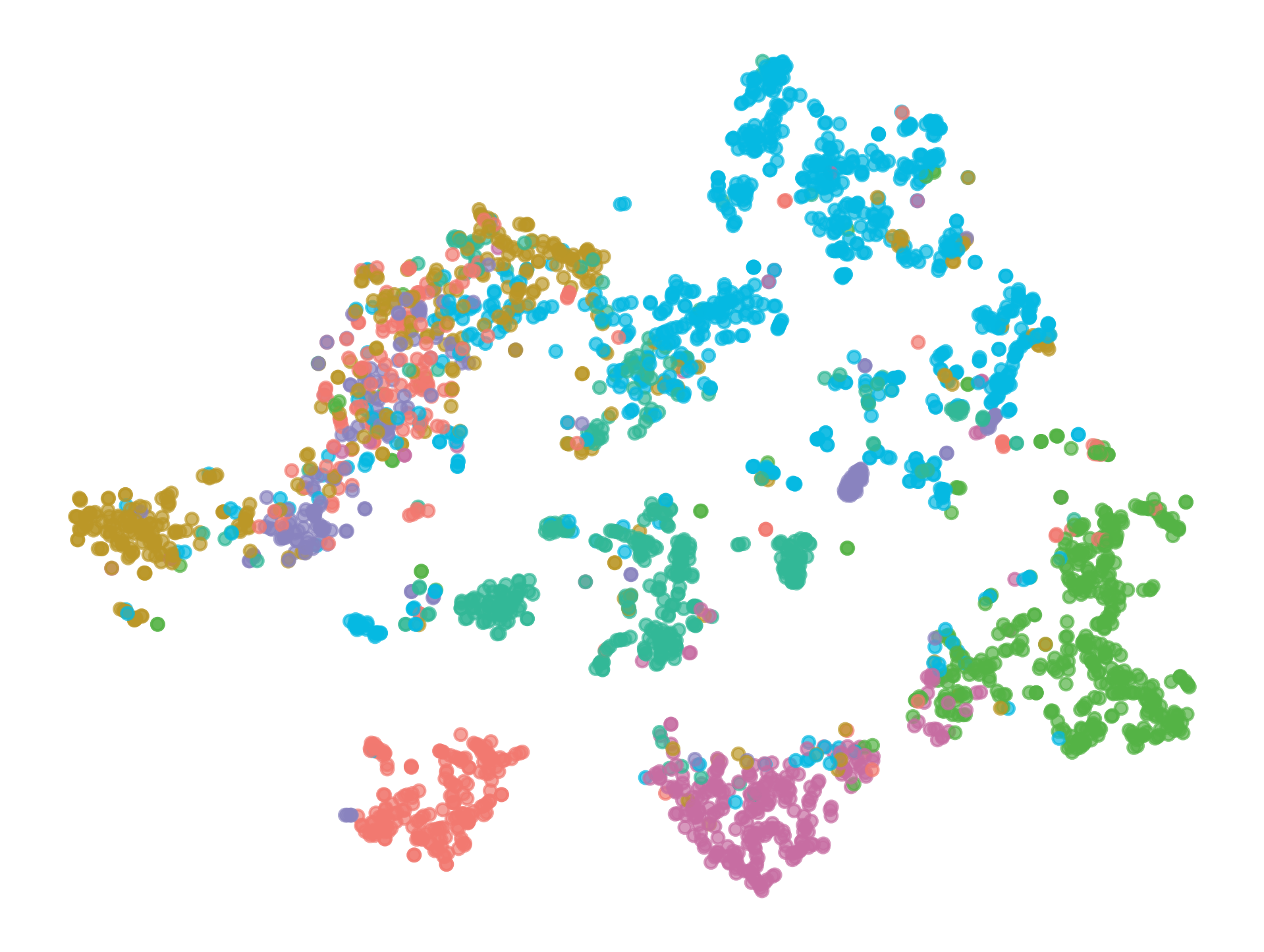}}
\vspace{-2mm}
\caption{Visualization results on CoraML.}
\vspace{-4mm}
\label{fig:visualization}
\end{figure*}
\subsubsection{Visualization}
To intuitively demonstrate the representation quality of S2-DiGCL compared with other unsupervised baselines, we visualize the learned node embeddings from the CoraML using t-SNE~\cite{van2008visualizing}.
The results are shown in Fig.~\ref{fig:visualization}, where different colors correspond to different node classes.

The embeddings produced by GraphCL appear widely scattered, with blurred class boundaries and significant overlap between clusters.
By comparison, DGI, UGCL, and DiGCL yield more compact and coherent clusters, indicating that these models better capture local structural patterns.
Nevertheless, both DGI and UGCL still display partially overlapping regions, suggesting that the learned representations are not fully discriminative in the latent space.
In contrast, S2-DiGCL produces clearly separated and compact clusters, where nodes of the same class are closely grouped while different classes occupy distinct regions in the embedding space.
We treat the visualization here as supportive qualitative evidence, which exhibit consistency with the quantitative improvements reported in node classification and link prediction.

\begin{table}[t]
\centering
\caption{Ablation study on CoraML and Citeseer.}
\vspace{-0.2cm}
\label{tab:ablation}
\resizebox{\linewidth}{!}{
\begin{tabular}{ccccc}
\toprule
\multirow{2}{*}{Variant} & \multicolumn{2}{c}{CoraML} & \multicolumn{2}{c}{Citeseer} \\
\cmidrule(r){2-3} \cmidrule(l){4-5}
& Node & Link & Node & Link \\
\midrule
\textit{w/o complex} & 71.41$\pm$1.48 & 82.29$\pm$2.10 & 62.68$\pm$1.49 & 76.54$\pm$3.41 \\
\textit{w/o real} & 73.18$\pm$1.92 & 85.51$\pm$1.70 & 64.84$\pm$1.48 & 82.56$\pm$2.46 \\
\midrule
\textit{w/o perturb} & 69.81$\pm$2.12 & 84.21$\pm$2.15 & 61.17$\pm$0.57 & 79.29$\pm$1.42 \\
\textit{uniform perturb} & 70.68$\pm$2.98 & 82.48$\pm$1.57 & 61.98$\pm$1.47 & 80.46$\pm$2.87 \\
\midrule
\textit{w/o direction} & 71.50$\pm$1.87 & 83.47$\pm$1.23 & 62.04$\pm$0.74 & 76.76$\pm$2.45 \\
\textit{w/o RNN} & 70.42$\pm$1.44 & 84.09$\pm$0.80 & 61.18$\pm$2.69 & 78.07$\pm$1.21 \\
\midrule
S2-DiGCL & \textbf{81.83$\pm$1.37} & \textbf{89.80$\pm$0.75} & \textbf{67.68$\pm$0.63} & \textbf{86.37$\pm$1.69} \\
\bottomrule
\vspace{-5mm}
\end{tabular}}
\end{table}

\subsection{Ablation Study}
\label{sec:ablation}
To address \textbf{Q2}, we perform an ablation study on the CoraML and Citeseer datasets to examine the contribution of each component within S2-DiGCL.
The analysis focuses on three key aspects: augmentation perspectives, personalized perturbations, and path augmentation.
Table~\ref{tab:ablation} reports the results for both node classification and link direction prediction.
We choose these two datasets because they provide stable evaluation for both tasks while covering different sparsity and homophily characteristics.

\subsubsection{Augmentation Perspectives}
We begin by disabling either the real-domain or complex-domain perturbation to analyze their individual effects.
The results show that removing either component leads to a noticeable drop in accuracy across both tasks, confirming that the two spatial perspectives contribute complementary information.
In particular, removing the complex-domain augmentation results in a more pronounced performance decline compared with removing the real-domain component.
This observation suggests that the personalized magnetic Laplacian perturbation in the complex domain contributes a stronger global structural signal, whereas the real-domain branch provides additional local directional cues that further improve the final representation.

\subsubsection{Personalized Perturbations}
Next, we evaluate the role of the personalized magnetic Laplacian perturbation by comparing two simplified variants.
The \textit{w/o perturb} and \textit{uniform perturb} settings correspond to fixing the personalization coefficient $r$ in Eq.~(\ref{eq:phi}) to 0.0 and 1.0, respectively.
Both variants lead to a significant reduction in performance, indicating that the adaptive perturbation mechanism is essential for capturing diverse local contexts.
The results support the usefulness of adaptive perturbation strengths for preserving structural nuances, especially in graphs exhibiting high heterogeneity or varying local connectivity.

\subsubsection{Path Augmentation}
Finally, we investigate the effect of the path sampler and the design of the real-domain encoder.
In the \textit{w/o direction} variant, the model ignores edge direction during path sampling, which weakens its ability to model asymmetric neighborhood structures.
This limitation particularly affects the understanding of directional relationships between central nodes and their neighbors.
Similarly, replacing the RNN-based encoder with an MLP in the \textit{w/o RNN} variant removes the sequential modeling capability required to encode ordered path information.
Both variants exhibit notable performance degradation compared with the complete S2-DiGCL model.
These findings highlight that directionality and sequence modeling play indispensable roles in capturing hierarchical and sequential dependencies within digraphs.

\begin{figure}
\centering
\subfigure[Perturb ratio]{\label{fig:param_ratio}\includegraphics[width=0.48\linewidth]{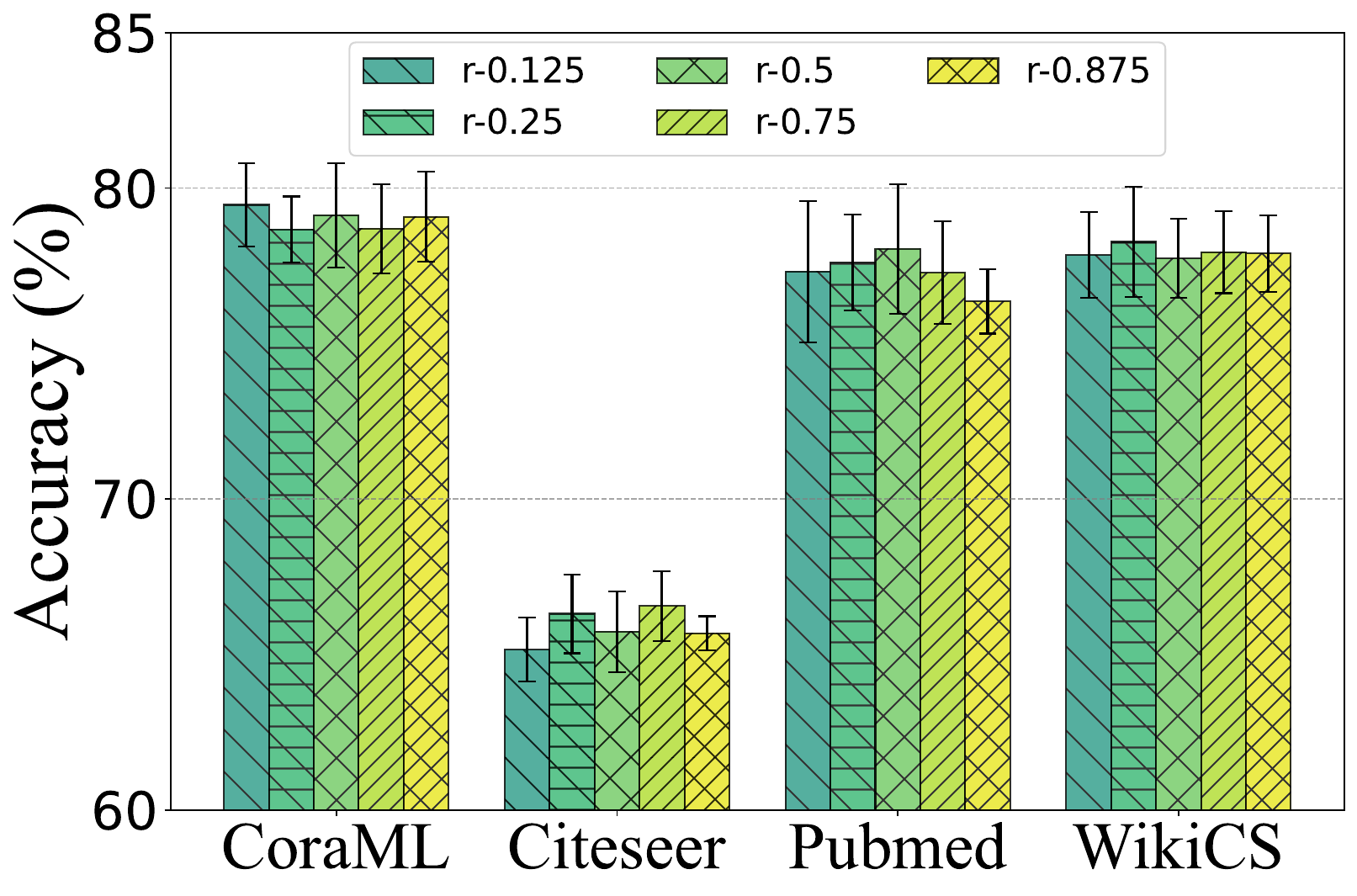}}
\subfigure[Sample length]{\label{fig:param_len}\includegraphics[width=0.48\linewidth]{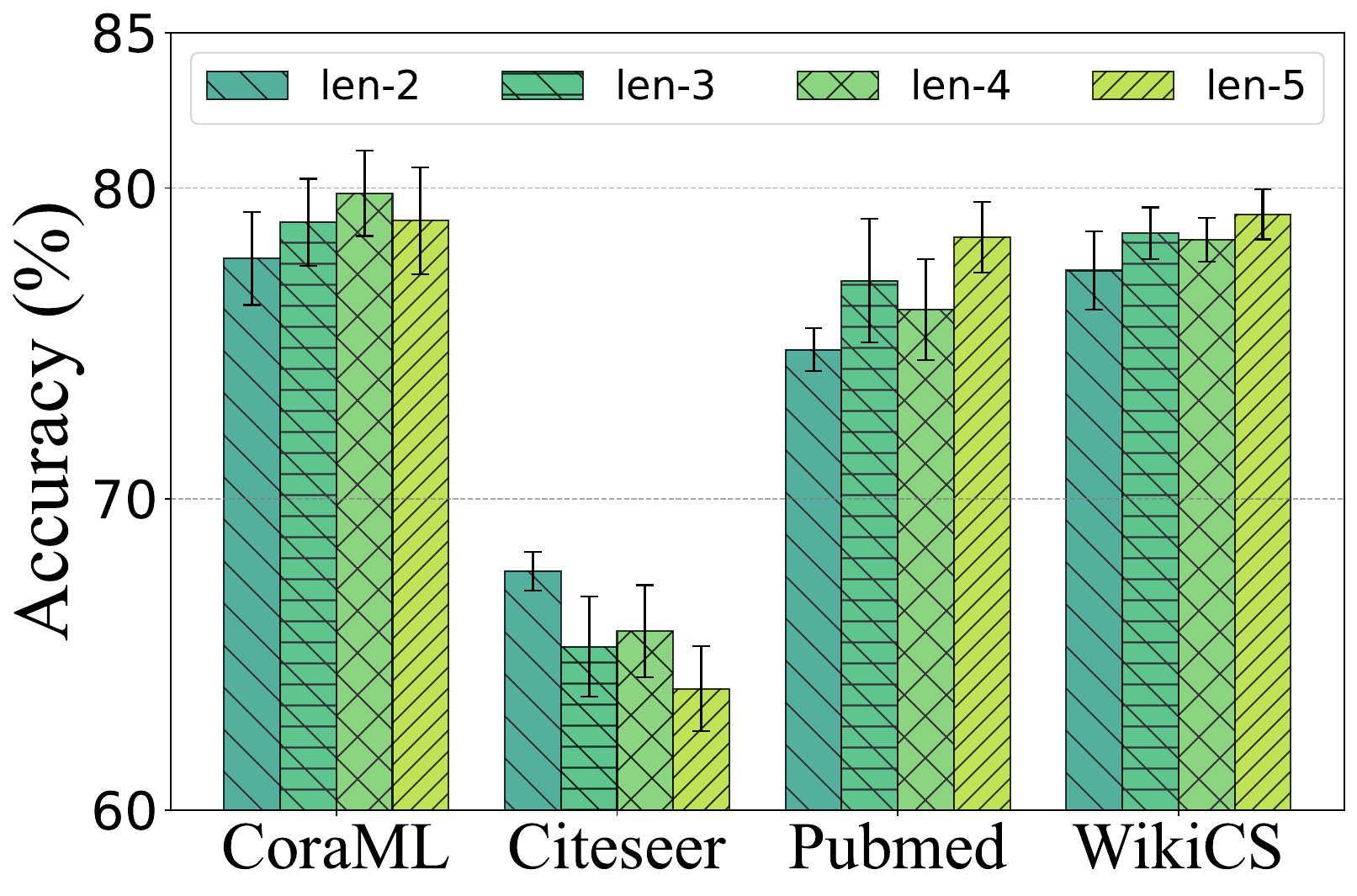}}
\vspace{-2mm}
\caption{Performance with different parameters.}
\vspace{-4mm}
\label{fig:param}
\end{figure}

\subsection{Sensitivity Analysis}
To address \textbf{Q3}, we perform sensitivity experiments to examine the influence of key hyperparameters on the performance of S2-DiGCL in the node classification task.
The analysis is conducted on four representative datasets: CoraML, Citeseer, Pubmed, and WikiCS.
We focus on two critical parameters that directly affect the dual-domain learning process:
\ding{192} the perturbation ratio $r$ within the personalized magnetic Laplacian, and
\ding{193} the path length $l$ within the direction-aware path sampler.
The corresponding results are illustrated in Fig.~\ref{fig:param}.

\subsubsection{Perturbation Ratio}
We vary the perturbation ratio $r$ across the range [0.125, 0.25, 0.5, 0.75, 0.875] to assess how the intensity of magnetic perturbation influences representation learning.
As shown in Fig.~\ref{fig:param_ratio}, the performance of S2-DiGCL remains consistently stable across all datasets, with only minor fluctuations.
This observation suggests that the personalized magnetic Laplacian perturbation effectively balances structural smoothness and local adaptability, enabling the model to capture both global and local topological information.
The robustness to variations in $r$ further indicates that S2-DiGCL does not rely on fine-tuned parameter settings, which enhances its practicality across diverse digraph scenarios.

\subsubsection{Path Length}
We also analyze the impact of the path length $l$ in the direction-aware path sampler by varying it among [2, 3, 4, 5].
The trends depicted in Fig.~\ref{fig:param_len} show that the optimal value of $l$ differs across datasets, with CoraML achieving its peak performance at $l = 4$, Citeseer at $l = 2$, and both Pubmed and WikiCS at $l = 5$.
These results reveal a trade-off between capturing sufficient directional information and avoiding noise propagation.
Shorter paths may fail to model higher-order directional dependencies, leading to incomplete neighborhood representations, while excessively long paths may accumulate redundant or noisy information, particularly in sparse graphs such as Citeseer.
Overall, the consistent performance across a moderate range of $l$ values demonstrates that S2-DiGCL achieves a good balance between expressiveness and stability in modeling digraph structures.

\subsection{Efficiency Analysis}
To address \textbf{Q4}, we evaluate the computational efficiency of S2-DiGCL with respect to convergence speed and total training time, comparing it against all unsupervised baselines on Citeseer.
From a complexity perspective, the main additional costs of S2-DiGCL come from three sources: constructing personalized magnetic perturbations over directed edges, sparse message passing on the magnetic Laplacian, and path sampling/sequence encoding in the real-domain branch.
Ignoring constant factors, the perturbation construction is linear in the number of directed edges, the complex-domain propagation scales with the number of layers and sparse edge interactions, and the real-domain branch further scales with the sampled path length and sequence encoder cost.
Therefore, S2-DiGCL remains practical on small-to-medium benchmark digraphs, while its scalability to substantially larger graphs still depends on more efficient sampling and propagation schemes.
The experimental results are presented in Fig.~\ref{fig:time}.

\subsubsection{Convergence Speed}
To evaluate convergence behavior, we monitor the classification accuracy of each model over training time.
As shown in Fig.~\ref{fig:conv_speed}, S2-DiGCL converges within approximately 15 seconds while maintaining a steadily increasing accuracy throughout the training process.
This convergence pattern indicates that the dual spatial learning strategy facilitates more efficient optimization by promoting rapid alignment between the real and complex representation spaces.
Although methods such as DGI, GCA, and GRACE reach convergence slightly earlier, their final accuracies are notably lower.
These results suggest that S2-DiGCL achieves a reasonable balance between convergence speed and representation quality on the evaluated benchmark.

\subsubsection{Training Time}
\begin{figure}
\centering
\subfigure[Convergence speed]{\label{fig:conv_speed}\includegraphics[width=0.48\linewidth]{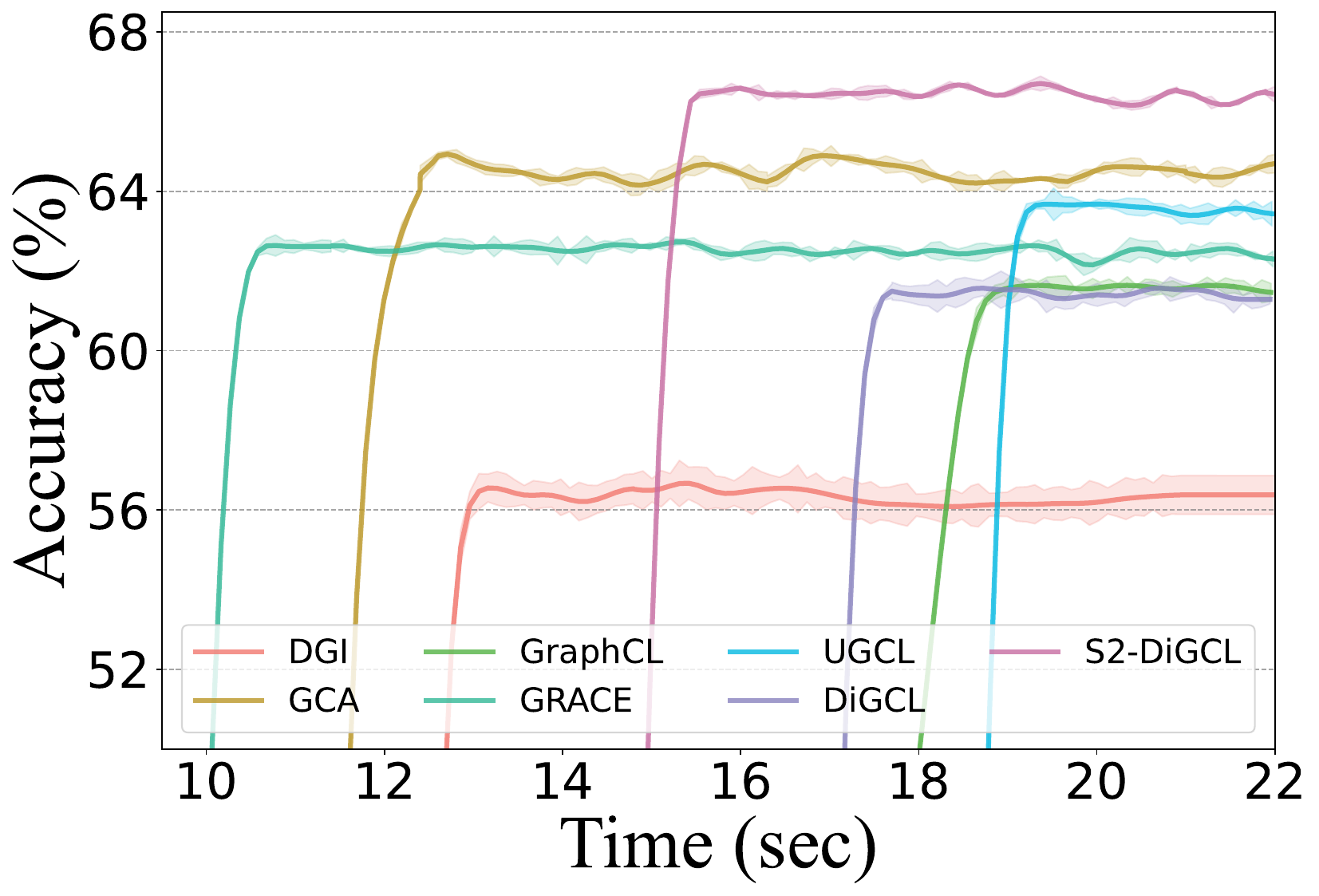}}
\subfigure[Training time]{\label{fig:train_time}\includegraphics[width=0.48\linewidth]{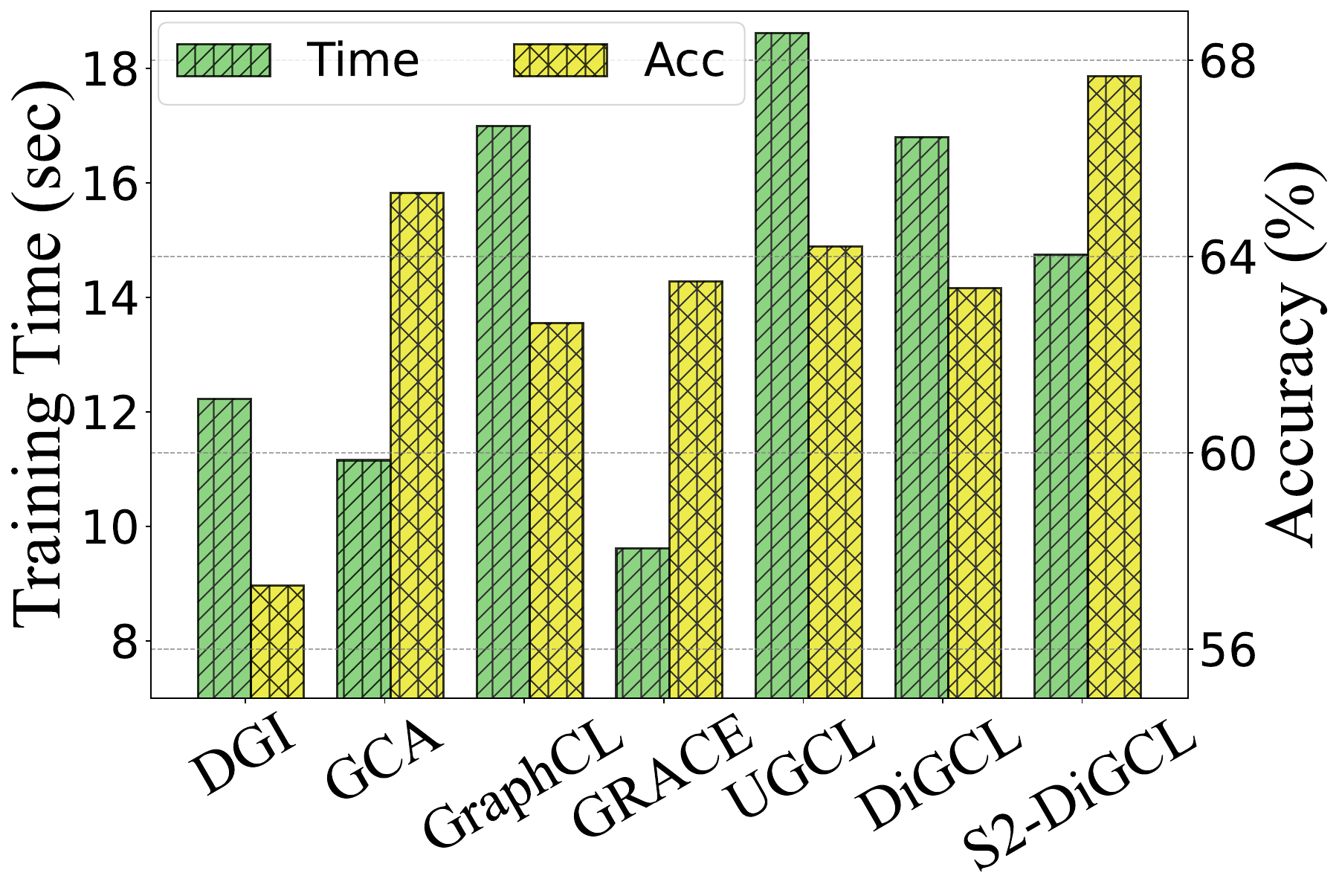}}
\vspace{-2mm}
\caption{Time efficiency on Citeseer.}
\vspace{-4mm}
\label{fig:time}
\end{figure}

We further compare the total training time and final accuracy of all models, as summarized in Fig.~\ref{fig:train_time}.
S2-DiGCL achieves the highest accuracy among all unsupervised methods while maintaining competitive training time.
This outcome indicates that the proposed dual-domain design provides favorable accuracy-efficiency trade-offs on Citeseer, although it still introduces additional overhead relative to simpler single-branch methods.
The efficient optimization behavior can be attributed to the complementary nature of the real-domain path augmentation and the complex-domain magnetic perturbation, which together enable faster convergence and more stable gradient propagation.
Overall, these results show that S2-DiGCL attains high accuracy with competitive runtime on the evaluated benchmark, while further improvements are still needed for substantially larger directed graphs.

\section{Conclusion}
In this paper, we introduced S2-DiGCL, a framework for contrastive learning on digraphs that unifies the strengths of complex- and real-domain representations.
The proposed dual spatial perspective effectively addresses the challenge of modeling inherent directionality in GCL.
By incorporating personalized perturbations within the magnetic Laplacian, S2-DiGCL captures global directional dependencies, while its path-based subgraph augmentation strategy preserves local asymmetric information flows.
Experiments on seven real-world datasets verify that S2-DiGCL consistently improves node classification performance over strong unsupervised baselines and achieves competitive link prediction results on four benchmark digraphs.
These results highlight the framework’s capability in learning direction-aware graph representations without label supervision.
Although the current augmentation process introduces additional computational cost, the empirical results indicate that the proposed dual-domain design provides a useful trade-off between accuracy and efficiency on benchmark-scale directed graphs.
Future work will focus on more scalable sampling and perturbation mechanisms, stronger cross-view fusion strategies, and broader evaluation on additional downstream tasks and larger digraph datasets.

\bibliographystyle{IEEEtran}
\bibliography{reference}

\end{document}